# CoNOAir: A Neural Operator for Forecasting Carbon Monoxide Evolution in Cities

Sanchit Bedi[1], Karn Tiwari[2], Prathosh A. P.[2,*], Sri Harsha Kota[1,*], N. M. Anoop Krishnan[1,3,*]

[1]Civil Engineering Department, Indian Institute of Technology Delhi, New Delhi, India 110016

[2]Electrical Communications Department, Indian Institute of Science Bengaluru, Bengaluru, India 560012

[3]Yardi School of Artificial Intelligence, Indian Institute of Technology Delhi, New Delhi, India 110016

[*]Corresponding authors: PAP (prathosh@iisc.ac.in), SHK (harshakota@iitd.ac.in), NMAK (krishnan@iitd.ac.in)

**Abstract**

Carbon Monoxide (CO) is a dominant pollutant in urban areas due to the energy generation from fossil fuels for industry, automobile, and domestic requirements. Forecasting the evolution of CO in real-time can enable the deployment of effective early warning systems and intervention strategies. However, the computational cost associated with the physics and chemistry-based simulation makes it prohibitive to implement such a model at the city and country scale. To address this challenge, here, we present a machine learning model based on neural operator, namely, Complex Neural Operator for Air Quality (CoNOAir), that can effectively forecast CO concentrations. We demonstrate this by developing a country-level model for short-term (hourly) and long-term (72-hour) forecasts of CO concentrations. Our model outperforms state-of-the-art models such as Fourier neural operators (FNO) and provides reliable predictions for both short and long-term forecasts. We further analyse the capability of the model to capture extreme events and generate forecasts in urban cities in India. Interestingly, we observe that the model predicts the next hour CO concentrations with $R^2$ values greater than 0.95 for all the cities considered. The deployment of such a model can greatly assist the governing bodies to provide early warning, plan intervention strategies, and develop effective strategies by considering several what-if scenarios. Altogether, the present approach could provide a fillip to real-time predictions of CO pollution in urban cities.

**Main**

Indian cities, epicentres of economic prosperity and cultural vibrancy, are poised to accommodate over 675 million residents by 2035 [1]. The rapid urbanization, alongside rampant industrialization and a burgeoning population, has triggered a notable surge in the concentration of airborne pollutants such as Carbon Monoxide (CO), a criteria pollutant extensively studied historically to gauge urban air quality [2-6]. While CO is recognized as an excellent tracer for atmospheric dynamics [7], it plays a pivotal role in tropospheric ozone chemistry [8] and exerts a profound impact on climate dynamics [9]. Furthermore, CO impacts neurodevelopment in children [10,11], and is associated with cardiovascular diseases [12,13].

Incorporating air quality management into policymaking for the development of sustainable urban centres is crucial [14-16]. Traditionally, data from ground-based analysers is utilized to prioritize air quality management scenarios [15]. However, a low-density monitor network fails to capture spatial variations; for instance, India has 544 ground-based continuous monitoring stations measuring CO [17], suggesting that the current state of air quality monitoring in Indian cities is inadequate [18,19]. This challenge can be addressed by supplementing ground-based observations with air quality models to enhance the spatial coverage. Several studies have compared the performance of air quality models such as Model for Ozone and Related Tracers (MOZART) [20-23] and Weather Research and Forecasting-Chem (WRF-Chem) [24-27] with ground-based observations and satellite retrievals for CO. Despite their widespread usage, the performance of these models ranges from poor to moderate, often resulting in underestimation of pollutant concentrations. Additionally, computations become increasingly expensive due to the involvement of the chemistry module and the enhanced spatial and temporal resolutions required [28]. These computations are performed for every instance with new initial boundary conditions, increasing the time required to make new simulations [29].

Data-driven machine learning (ML) models are gaining traction as substitutes for these conventional models, given the availability of large amounts of data for training. ML models offer the distinct advantage of rapidly generating inferences on new instances with significantly lower computational demands. Various ML algorithms such as Linear Regression, Artificial Neural Networks, K-Nearest Neighbour, Support Vector Regression, Long short-term memory, and Convolutional Neural Networks (CNNs) were employed by several authors in the past to model CO [30-40].

However, these ML models learn the function map between the finite-dimensional input and output spaces and need to be retrained for each new case i.e. a separate model for every new city. This issue can be mitigated by adopting the approach from conventional models—learning the infinite-dimensional operator between input and output function spaces using partial differential equations. Such a data-driven model would also address the constraints related to mesh-dependency and resolution invariance [41]. Some approaches in this direction include the work on Fourier Neural Operator (FNO) [42], which can learn the map between initial boundary conditions to the time-evolution of further steps. However, these models rely on stationary signals and do not perform well for modelling pollutants such as CO, which are inherently non-stationary.

To address the challenges of (a) expensive computations, (b) slow inference speeds, (c) inability of previous ML models to learn the underlying dynamics, and (d) issues with mesh-dependency and generalizability of the models, a neural operator architecture, namely Complex Neural Operator for Air Quality (CoNOAir) [43], to model the evolution of CO concentrations over the entire country is developed. CoNOAir employs a complex neural network along with the fractional Fourier transform that can capture the non-stationary signals, as in the case of CO concentrations. This model is trained using four years' hourly data from WRF-Chem simulations that work well for regional-scale air quality modelling. Furthermore, the performance of this model was compared to that of state-of-the-art neural operator FNO. The present framework can serve as a useful tool for generating real-time forecasts of CO concentrations in urban cities of a country in an extremely economical fashion, making it suitable for integration into urban planning.

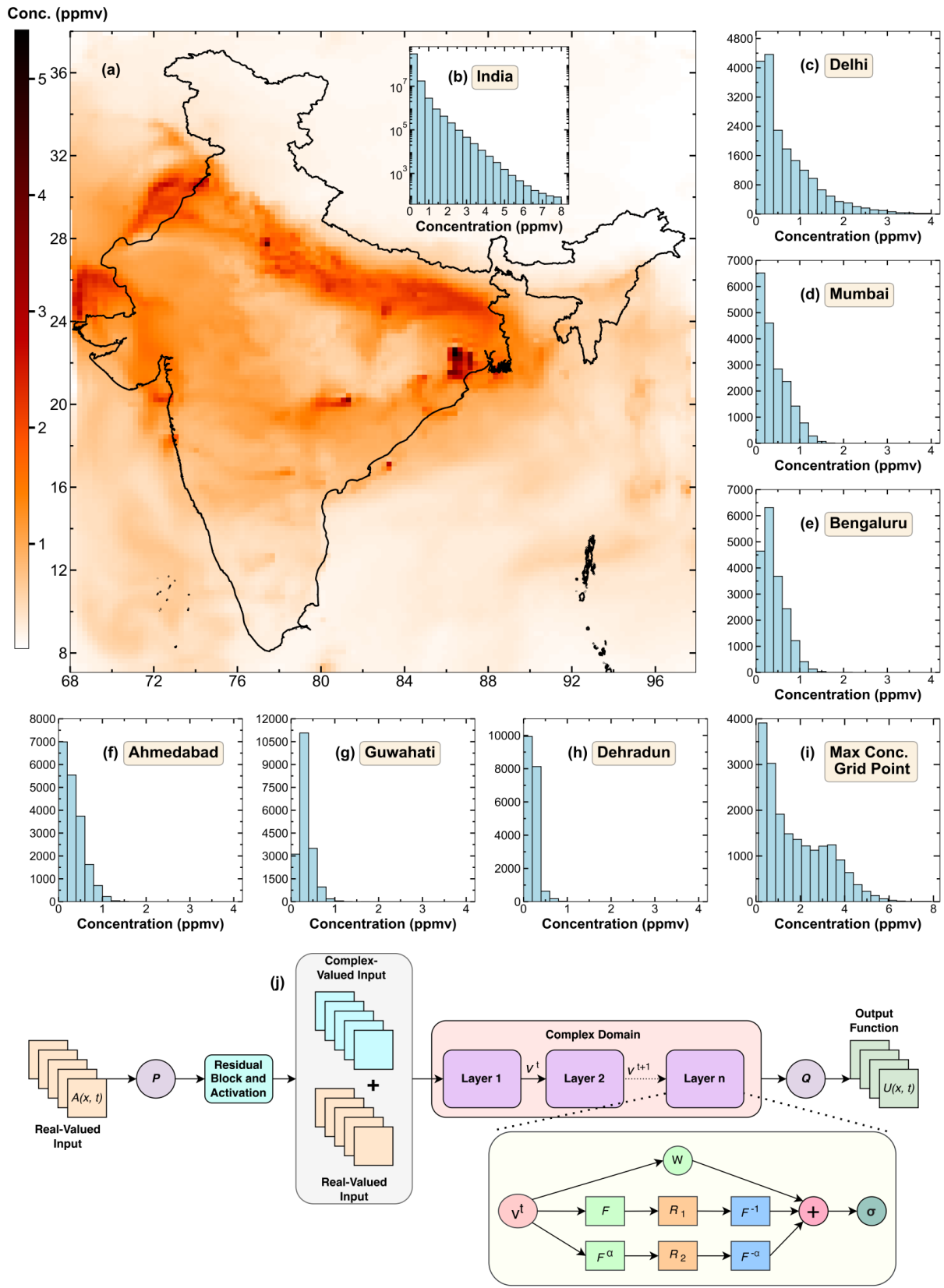

**Figure 1. Data visualisation and architecture.** *(a) Indian subcontinent for the timestamp of 03$^{rd}$ March 2016 at 04:00 AM IST having the maximum cumulative CO concentration. (b) Histogram of CO concentration over the entire subcontinent for the entire study period considered from 2016 to 2019. Histograms of CO concentration for the same timestamp in six*

*major cities, namely, (c) Delhi, (d) Mumbai, (e) Bengaluru, (f) Ahmedabad, (g) Guwahati, and (h) Dehradun, (i) Histogram of CO concentration corresponding to the maximum concentration grid point. (j) Proposed CoNOAir architecture: Time-varying input undergoing lifting operation followed by kernel integration through FrFT layers (computations inside the layer zoomed in) and projected back to lower dimension to generate output.*

**Results**

*CO concentration in Indian cities*

In this work, we considered several major cities of India by modelling the CO evolution at the country level for a period from 2016 to 2019. Note that the CO evolution for this period was simulated using the WRF-Chem model (v3.9.1), henceforth referred to as ground truth (for details, please refer to the Methods section). Figure 1(a) depicts the study area spanning from 7.060º N to 38.004º N and 67.791º E to 97.934º E with the CO concentrations. Note the timestamp (03$^{rd}$ March 2016 at 04:00 AM IST) corresponding to the maximum cumulative CO concentration is used, representing the extreme air pollution event over the entire study area. We have also identified the grid point having the maximum concentration at this timestamp, which is given by the coordinates 23.316º N, 86.416º E. This grid's proximity to several thermal power plants and industrial areas could be a cause of elevated concentrations. Figure 1 (b) shows the distribution of CO concentrations observed over all the grid points and over all the time stamps for the years 2016, 2017, 2018 and 2019.

For analysing urban CO concentration, India is divided into six major zones to group the airsheds covering the 131 cities under the National Clean Air Programme (NCAP) namely South, Central, Northeast, Indo-Gangetic Plain (IGP), Northwest, and Himalayan [44]. Although these airsheds have been defined based on Particulate Matter (PM), we use the same zones for CO pollution as well and have identified six major cities in these zones for detailed analysis. The identified cities are designated as non-attainment cities in the NCAP plan developed by the Indian government and are targeted for reducing their PM$_{2.5}$ concentrations. Each city represents one of the six airsheds and is the largest city belonging to that airshed. The cities are namely Delhi (28.602º N, 77.103º E) in IGP, Mumbai (18.975º N, 72.937º E) in Central, Ahmedabad (23.091º N, 72.692º E) in Northwest, Bengaluru (13.089º N, 77.593º E) in South, Dehradun (30.309º N, 78.084º E) in Himalayan, and Guwahati (26.209º N, 91.808º E) in Northeast. Figure 1 (c), (d), (e), (f), (g), and (h) depict the distribution of CO concentration over these cities for the entire simulated data. In comparison to the 'good' air quality index thresholds for CO [45], Delhi transgressed the threshold 30.22% of the time, followed by Mumbai (10.5%), followed by Bengaluru (6.7%) and Ahmedabad (3.5%). While Guwahati and Dehradun exceeded the thresholds only 0.71% and 0.05% of the time, respectively, the grid point registering the highest concentration surpassed the limit in 63.86% of instances. Figure 1 (i) depicts the distribution of CO concentrations over the maximum concentration grid point.

*Complex Neural Operator for Air Quality*

To predict the evolution of CO, we present our neural operator-based framework namely, Complex Neural Operator for Air Quality (CoNOAir). Figure 1(j) shows the CoNOAir architecture for forecasting CO. The model uses the CO concentration from the previous $k$ timesteps to forecast the concentrations in the next timestep, which is then used in an autoregressive fashion to forecast the CO concentration in further timesteps. Here, we consider $k$ as 10 timesteps. The model consists of three primary operations, namely, lifting, kernel integration, and projection. First, the model projects input to a high-dimensional latent space using a fully dense pointwise neural network. Then, the model employs a series of affine operations along with kernel integration, that are passed through a non-linear activation function. Finally, the output is obtained by projecting the final embedding in the latent space through a dense point-wise neural network. In contrast, to classical neural operators such as FNO, CoNOAir employs a fractional Fourier transform for the kernel integration along with complex neural networks for the affine transformations. As demonstrated later, these features are particularly effective for capturing non-stationary signals such as the CO evolution. The models are trained by minimizing the Relative $L^2$ norm ($RL^2$ Norm) loss of the predicted CO concentration over *n* future timesteps. Here, we present three models with *n* values as 4, 6, and 16. For comparison, we also train FNO models under the same setting. Note that the models are trained on the data from 2016 to 2018 and the data from 2019 is kept as the test set for evaluating the model. Further details of the models and the hyperparameters are provided in the Methods section.

*Forecasting CO concentrations*

First, we evaluate the performance of CoNOAir in comparison with FNO for forecasting CO concentrations. Specifically, we consider three models trained on 4, 6, and 16 forward timesteps as outlined in the Methods section (refer to Figure S1 for the train (a) and test (b) loss curves). To evaluate the performance of these models, we compute the error metrics for three different cases: 16, 20, and 72 hours forecasting.

Table 1 shows the performance of CoNOAir and FNO models, trained on 4, 6, and 16 forward timesteps, for 16, 20, and 72 hours forecasting. CoNOAir outperforms FNO in all the cases for a similar setting. Interestingly, while the CoNOAir(4) performs best for 1-hour forecast, CoNOAir(16) performs best for longer-term forecast of 16 hours. This is due to the fact that CoNOAir(16), being trained on 16 forward steps, exhibits better longer-term stability in predictions, whereas CoNOAir(4) exhibits better short-term predictions with the errors accumulating for the long-term predictions. Similar performance is observed for $20^{th}$ and $72^{nd}$ hour predictions, with CoNOAir(16) giving the best results in both cases. More detailed discussions on these, along with the plots, are included in the Supplementary Material in Section S.2. Altogether, it is evident that a given variation of the CoNOAir model performs better than the corresponding variation of FNO model. Moreover, while CoNOAir(16) give the best performance for forecasting 16, 20, and 72 hours, CoNOAir(4) gives the best performance for the next hour prediction. Thus, depending on the fidelity and timescale of interest, the respective CoNOAir models could be used for forecasting the CO concentrations. Since, CoNOAir consistently outperforms FNO both in short-term and long-term forecasts, henceforth, our discussion is focused only on CoNOAir with the rest of the figures being included in Supplementary Material.

**Table 1. RMSE, MAE, RL² Norm Error for the evaluation of different datasets by the 6 models trained in this study.** The values in each cell depict the mean (standard deviation) obtained over all the samples over which the errors are calculated. The first part of the table represents the errors over the validation data, followed by the errors over the test data. The third part of the table shows the high concentration days from test data over which the long-term evaluation (72 hours) is performed.

| Error → Model ↓ | FNO(4) | FNO(6) | FNO(16) | CoNOAir(4) | CoNOAir(6) | CoNOAir(16) |
|---|---|---|---|---|---|---|
| Validation Data (Errors for 16$^{th}$ hour forecasts) | | | | | | |
| RMSE | 0.064 (0.041) | 0.056 (0.034) | 0.043 (0.022) | 0.058 (0.036) | 0.047 (0.028) | 0.031 (0.017) |
| MAE | 0.034 (0.02) | 0.03 (0.016) | 0.024 (0.012) | 0.029 (0.017) | 0.025 (0.014) | 0.018 (0.009) |
| RL² Norm | 0.284 (0.099) | 0.258 (0.088) | 0.195 (0.057) | 0.247 (0.083) | 0.205 (0.067) | 0.138 (0.042) |
| Test Data (Errors for 20$^{th}$ hour forecasts) | | | | | | |
| RMSE | 0.084 (0.046) | 0.078 (0.044) | 0.073 (0.031) | 0.084 (0.039) | 0.074 (0.031) | 0.072 (0.03) |
| MAE | 0.041 (0.017) | 0.036 (0.014) | 0.034 (0.009) | 0.038 (0.011) | 0.034 (0.009) | 0.032 (0.009) |
| RL² Norm | 0.382 (0.146) | 0.371 (0.169) | 0.343 (0.135) | 0.377 (0.142) | 0.339 (0.122) | 0.334 (0.125) |
| Elevated Pollution Days Subset from Test Data (Errors for 72$^{nd}$ hour forecasts) | | | | | | |
| RMSE | 0.145 (0.046) | 0.104 (0.027) | 0.09 (0.027) | 0.115 (0.03) | 0.111 (0.032) | 0.082 (0.024) |
| MAE | 0.089 (0.02) | 0.063 (0.012) | 0.053 (0.013) | 0.066 (0.012) | 0.064 (0.013) | 0.047 (0.01) |
| RL² Norm | 0.468 (0.073) | 0.415 (0.086) | 0.33 (0.055) | 0.419 (0.1) | 0.395 (0.15) | 0.3 (0.063) |

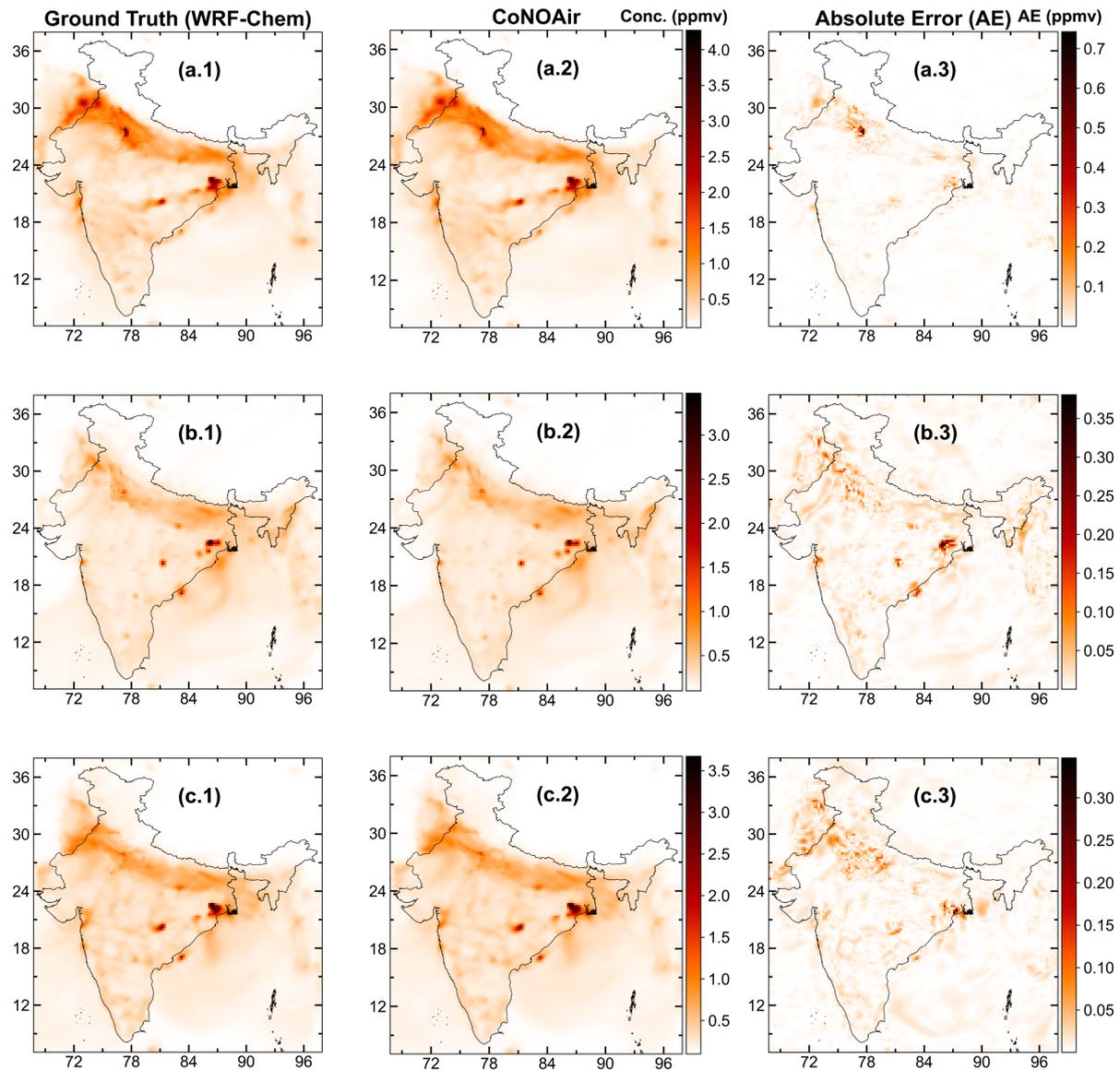

**Figure 2. Prediction of CoNOAir(4) on the Maximum Concentration Hours.** *Three maximum concentration hours are identified to assess the performance of CoNOAir. Both the ground truth and CoNOAir maps are normalized to the same range for the purpose of visualisation. Ground Truth for 12th February 2019 07:00 am IST (a.1), 6th February 19:00 IST (b.1) and 6th February 00:00 IST (c.1). CO concentrations predicted using CoNOAir(4) for the corresponding hours (a.2), (b.2) and (c.2). Absolute error between the ground truth and CoNOAir(4) predictions (a.3), (b.3) and (c.3).*

*Predicting Extreme Pollution Event*

Predicting extreme events accurately is important for developing early warning systems in cities. Moreover, since ML models are statistical in nature and most of the training data points are generally closer to the mean, these models may not perform well on extrema. Figure 2 shows the ability of CoNOAir(4) to predict the three worst-case scenarios, namely, 12th February 2019, 07:00 am IST, 6th February 2019, 19:00 IST and the midnight hour on 6th February 2019 IST. These three timestamps had the highest summed-up concentrations over the entire test data for the corresponding hour of the day, and represent morning rush hour, evening rush hour, and night-time radiation inversion (which restricts the potential dispersion of air pollutants), respectively. Figure 2 (a.1), (a.2), (a.3) show the ground truth, CoNOAir output and Absolute Error between these two for 12$^{th}$ February 2019, 07:00 hours IST. Similarly, Figure 2 (b) and (c) panels represent the same quantities for 6$^{th}$ February 2019, 19:00 hours IST and 6$^{th}$ February 2019, 00:00 hours IST, respectively. We observe that CoNOAir is able to capture the hotspots (grids of high concentrations) and the overall dispersion pattern of CO over the study area very well for all three scenarios. For both the second and third cases, the maximum absolute error is around 10%, while for the first scenario, it is around 17%. The efficiency of the model to learn the trend and identify the hotspots make them suitable for use in policy formulations. Figure S3 depicts that the FNO(4) configuration has more grids having high absolute errors, evident by a larger spread of dark regions along with a greater maximum absolute error than CoNOAir(4).

*Short-Term Forecasting for Six Cities*

Figure 3 shows the performance of CoNOAir(4) over the six cities listed earlier and the maximum concentration grid point for short-term forecasts. Specifically, we focus on the next hour forecasting capability for the month of February 2019 (from the test set) for these locations. Interestingly, we observe that CoNOAir(4) captures the fluctuations, including the spurious ones with sudden increases, in CO concentration both quantitatively and qualitatively. To demonstrate this further, we analyse the parity plot of predicted with respect to ground truth CO concentrations for these locations (see Figure 3). We observe that the $R^2$ values for all the locations are more than or equal to 0.95 suggesting a high accuracy in the prediction. From Figure S4, we can infer that the $R^2$ values for the seven grids for FNO(4) are less than or equal to CoNOAir(4). Tables S1 and S2 suggest that CoNOAir(4) has the best performance both in terms of RMSE and MAE for every city and the maximum concentration grid point. The RMSE and MAE, even for the CoNOAir(16), are comparable to FNO(4). These results reinforce the superior performance of CoNOAir both in terms of short-term and extreme value forecasting.

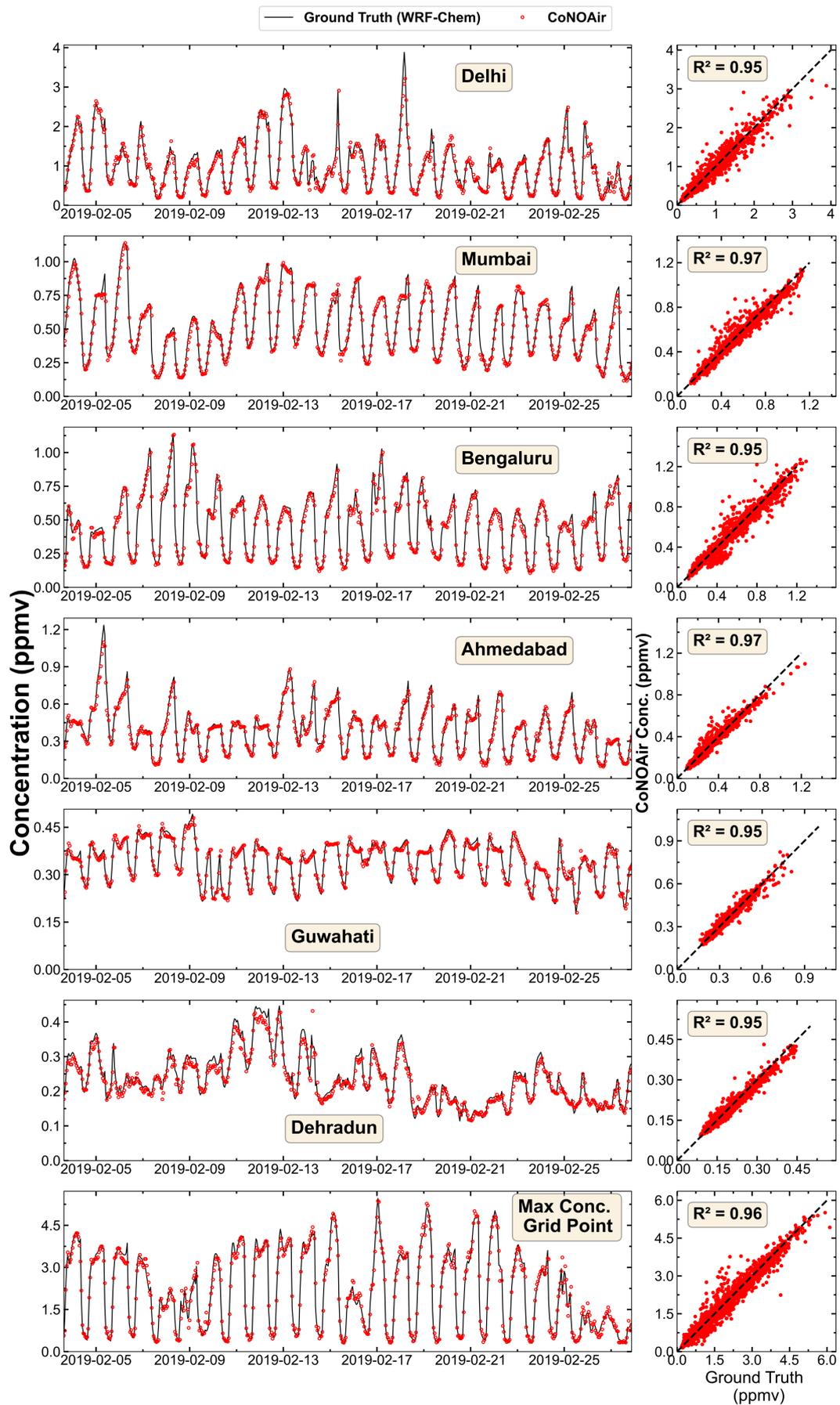

**Figure 3. Short-term forecast for cities.** *Time series 1-hour forecast for CoNOAir(4) for 6 cities and maximum concentration grid point to evaluate the ability of CoNOAir for grid-level short-term forecasts. A Time-series comparison is shown for February month in the left column. The column on the right shows the predicted values with respect to the ground truth for the whole test dataset. The $R^2$ values corresponding to each grid point are shown in the inset.*

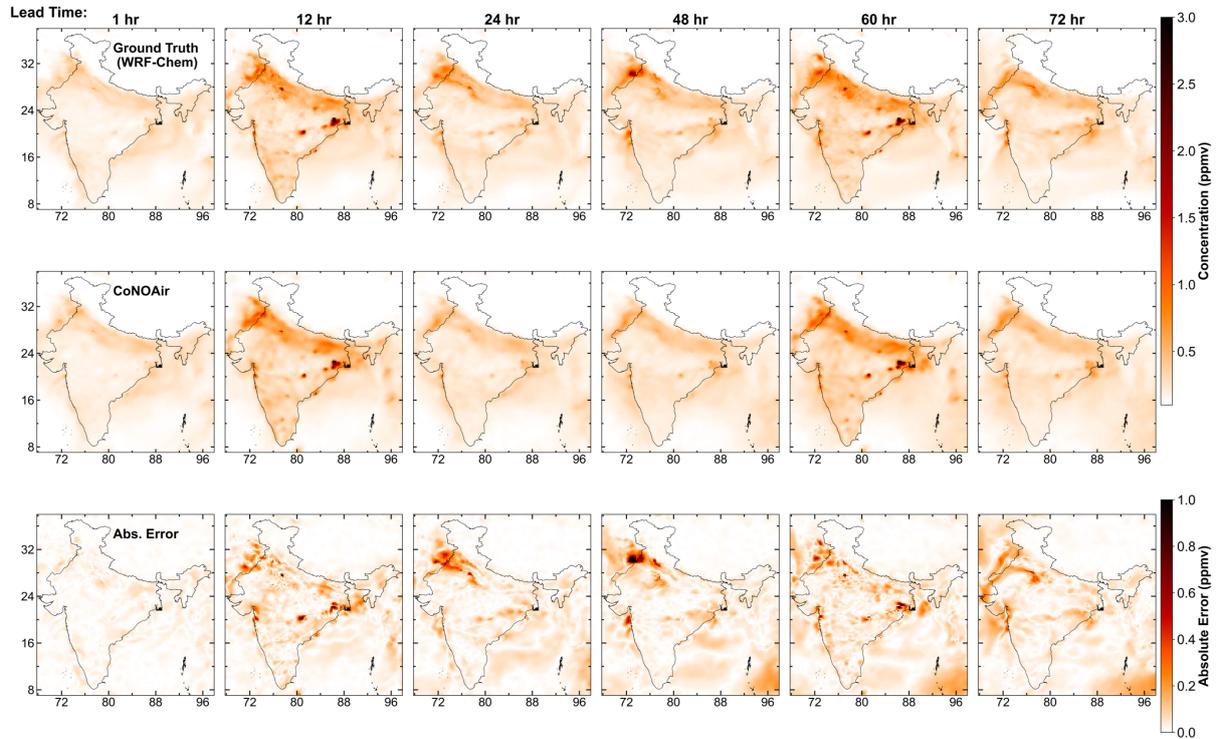

**Figure 4. Country-wide evolution of long-term forecast.** *72 hours autoregressive forecasts for one instance for the whole country for evaluating the long-term forecasting capability of the model. The top row represents the ground truth, while the middle row shows the CoNOAir(16) predictions. All the maps are normalized between 0 to 3 ppmv for the purpose of easy comparison. The absolute error between these two sets of plots is also shown in the bottom row which is normalized between 0 to 1 ppmv.*

*Long-Term Forecasting*

Long-term forecasting is an important aspect in pollution for strategic planning of the cities and policy making. We now analyze the capability of CoNOAir to perform reliable long-term forecast of CO concentration. Figure 4 shows one instance of 72 hours forecast using CoNOAir(16) starting 10th February 2019 at 12:00 IST in an autoregressive fashion (see Supplementary Materials for the results of the remaining models, Figures S5 to S9). The ground truth and the corresponding prediction using CoNOAir for the 1st, 12th, 24th, 48th, 60th and 72nd hour forecasts are shown for comparison (see Figure 4). We observe that CoNOAir can provide reasonable forecast both qualitatively and quantitatively for the 72nd hour as well. Specifically, CoNOAir is able to reproduce the overall patterns of dispersion and the distribution of the CO concentrations over the country with sufficient accuracy. On a closer examination, CoNOAir appears to be underestimating the concentrations over the country on longer timescales. The selection of this timestamp, i.e., noon hour when the potential of the dispersion of pollution is high, also explains the ability of our proposed model to build over relatively lower concentrations autoregressively to determine the future concentrations. The evolution of error metrics over all samples is given in Figures S2 (g), (h), and (i).

Finally, we analyse the long-term forecast of CoNOAir(16) for the selected cities and the maximum concentration location for 72 hours for the same instance as above (see Figure 5). CoNOAir(16) performs reasonably well for all the cities except Dehradun where the concentrations of CO are low. This shows the model's competence to efficiently replicate a greater range of concentration values autoregressively (Table S4 depicts the RMSE for forecasting 72 hours over these cities). It is impressive that the model's performance over the maximum concentration grid point is reasonably accurate even for a $72^{nd}$ hour forecast. This explains the superiority of the CoNOAir(16) model in forecasting CO levels at grids with high concentrations, that too autoregressively, making it a reliable model to be used for developing warning systems for predicting future episodes and preparing for them.

**Discussion**

Air quality modelling using ML algorithms can adeptly address challenges inherent in conventional models. This study advances ML models to simulate CO concentrations across India through operator learning. Neural Operator became our natural choice for this work since it has the advantage of approximating an infinite-dimensional PDE operator in finite-dimension space using only data, thus creating a generalized model that overcomes the issue of mesh-dependency inherent in CNN-based architectures. We trained the existing FNO model and introduced an enhanced architecture, CoNOAir, utilizing the fractional Fourier transform to better represent non-stationary signals and optimize parameters in complex space for learning the time evolution of the advection-diffusion equation. Training the FNO and CoNOAir models ranged from 4 to 80 hours on a single NVIDIA V100 GPU (32 GB memory). Further, generating inferences with these ML models took mere seconds to minutes for 72-hour forecasts, highlighting their potential to address computational time-related constraints of conventional models.

We evaluated the performances of three CoNOAir and FNO configurations based on short-term (1 hour) and long-term (3 days) forecasting. CoNOAir configurations consistently outperformed their FNO counterparts, indicating that FNO would require more parameters to achieve a similar level of performance. Detailed performance analysis over six major cities and the location with the highest CO concentrations revealed that CoNOAir(4) excelled in one-hour forecasts, while CoNOAir(16) outperformed or matched FNO(16) across all grids for forecasting 72 hours in an autoregressive fashion.

The super-resolution capability of the operators allows predictions at multiple resolutions for the micromanagement of hotspots and cities, facilitating the development of advanced intervention strategies. Future directions include integrating physics and chemistry-based information for precise CO and other pollutant predictions and developing real-time early warning systems. This framework could potentially be scaled globally to forecast CO concentrations for any city worldwide, offering enhanced management scenarios and improving living standards.

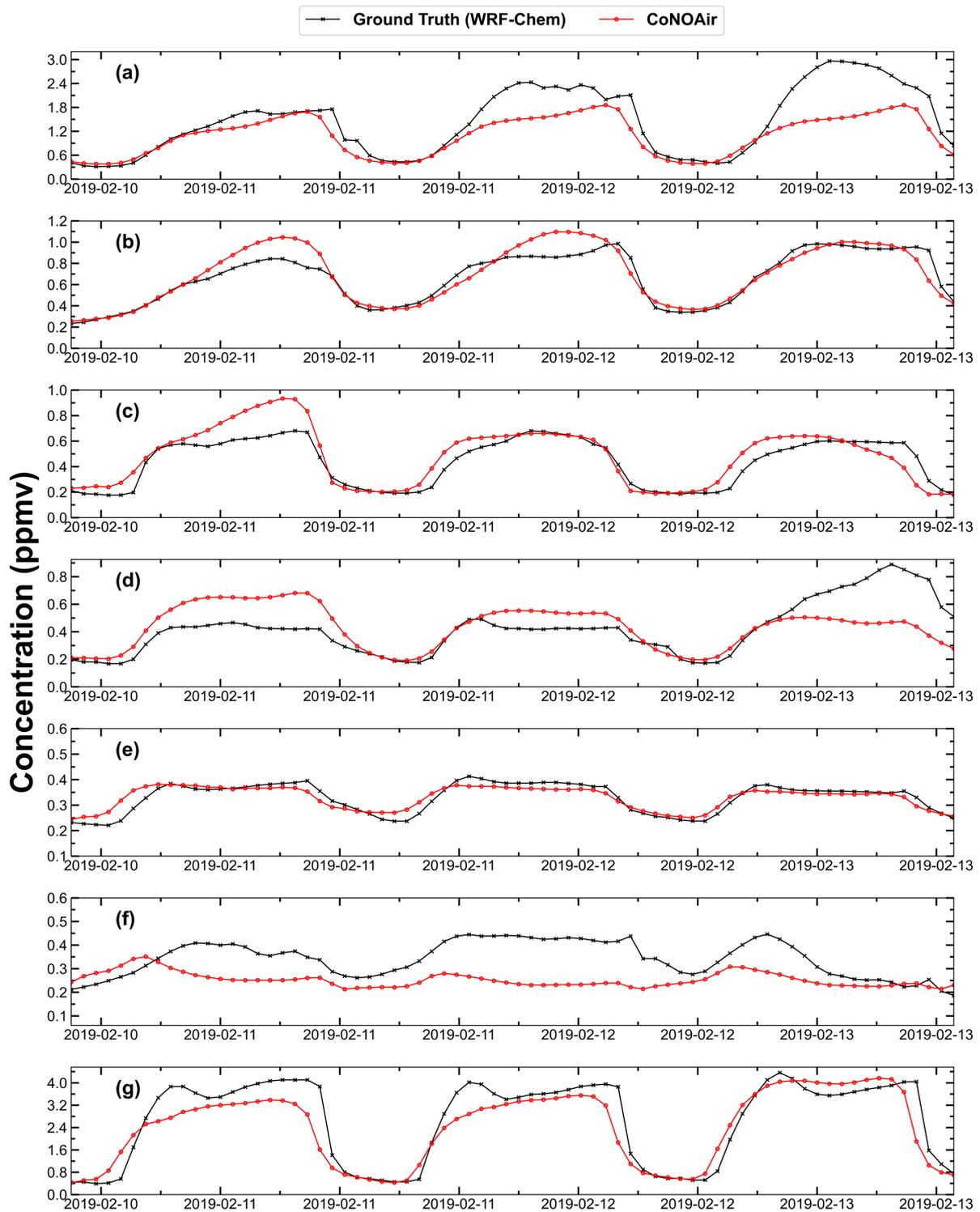

**Figure 5. Grid-level evolution of long-term forecast.** *72 hours forecasts for the same instance as Figure 4 using CoNOAir(16) for 6 cities (a) Delhi, (b) Mumbai, (c) Bengaluru, (d) Ahmedabad (d) Guwahati, (e) Dehradun and (f) Maximum concentration grid point for evaluating the long-term ability of the model to generate grid-level forecasts.*

## Methods

*Data Generation and Preprocessing*

In this study, we used the Weather Research and Forecasting coupled with the chemistry module (WRF-Chem) model (v3.9.1) for simulating air pollutant concentrations from 2016 to 2019 across India. More details about the configuration and validation of WRF-Chem is available elsewhere and is only briefly discussed here [25,46-50]. Simulations were performed at a spatial resolution of $25 \times 25$ km$^2$ comprising of $140 \times 124$ grids covering India. Model for Ozone and Related Chemical tracers' version 4 (MOZART-4) along with gas-phase chemistry coupled with Model for Simulating Aerosol Interactions and Chemistry (MOSAIC-4 bins) was employed. The boundary conditions for the chemical fields were taken from the Community Atmosphere Model with Chemistry (CAM-Chem) Global Chemical Transport Model at 6 h temporal resolution. The Emissions Database for Global Atmospheric Research (EDGARv5.0) was used for the anthropogenic emissions of various pollutants at a resolution of 0.1º. These emissions from 26 sectors were combined into 5 major sectors and scaling factors were applied on the base year 2015 for every sector for the subsequent years. The WRF-Chem output, generated in the form of a NetCDF file, was extracted as NumPy arrays and saved into *.mat file format for training the models. The ground-level CO concentrations as 2D arrays for the entire country with a temporal resolution of 1 hour (considered as ground truth in this study), and the corresponding timestamp were extracted. Subsequently, for the purpose of training and validation, respective datasets were prepared using the concentrations simulated for the years 2016, 2017, and 2018. To generate the training and validation set, we used the moving window approach to generate smaller time series each of 26 hours from the ground-truth data. All the samples were then shuffled randomly to obtain 3000 samples for training and 500 samples were obtained for testing the model's performance during the training as validation set. The simulated data for the year 2019 was kept separately as a hold out (test) set to assess our models' performances. The training, validation and test datasets were then normalized to the [0, 1] range using the absolute minimum and maximum values from the training dataset, considering both temporal and spatial dimensions to accelerate the training process.

*Model Architecture*

A neural operator, CoNOAir is developed in this study to model CO concentrations and the performances are evaluated against the ground truth. Further, we compare the performance of CoNOAir with one of the most widely used neural operators, namely, FNO, which gives state-of-the-art performance in several problems related to the time evolution of partial differential equations [42,51,52]. The problem formulation of the Neural Operator and its theory is presented subsequently.

Let $d \in \mathbb{N}$, $\Omega \in \mathbb{R}^d$ denotes bounded open, set with $A = A(\Omega; \mathbb{R}^{d_a})$ and $U = U(\Omega; \mathbb{R}^{d_u})$ denotes separable Banach spaces of functions, representing elements in $\mathbb{R}^{d_a}$ and $\mathbb{R}^{d_u}$, respectively. Suppose $G^\dagger: A \to U$ represents the underlying nonlinear surrogate mapping derived from a solution operator of parametric PDE. It is assumed that independent and identically distributed (i.i.d.) observations $(a_j, u_j)_{j=1}^N$ are available, where $a_j \sim \mu$ are sampled from the underlying probability measure $\mu$ supported on A, and $u_j = G^\dagger(a_j)$.

The aim of operator learning is to approximate $G^\dagger$ using a parametric mapping $G: A \times \Theta \to U$ or equivalently, $G_\theta: A \to U, \theta \to \Theta$, belongs to a finite-dimensional parameter space $\Theta$. The objective is to identify $\theta^\dagger \in \Theta$ such that $G(.,\theta^\dagger) = G_\theta^\dagger \approx G^\dagger$. It enables learning in infinite dimensional spaces by solving the optimisation problem in Eq. 1 constructed with a loss function $\mathcal{L}: U \times U \to \mathbb{R}$.

$$\min_{\theta \in \Theta} E_{a \sim u}[\mathcal{L}(G(a,\theta), G^{\dagger}(a))] \tag{1}$$

In neural operator learning, $G_\theta$ is parameterized using deep neural networks. is optimised using a data-driven approximation of the loss function. $G_\theta$ consists of three main components as follows:

a) Lifting Operation ($\mathcal{P}: \mathbb{R}^{d_a} \to \mathbb{R}^{v_0}$): projects input to high-dimensional latent space using a fully dense pointwise neural network along the channel dimension.

b) Kernel Integration Operation (K): represented iterative kernel operation that maps through series of linear operator and non-linear activation. It can be represented as follows:
$$V_{l+1} = \sigma(Wv_l + b_l + K(a;\phi)v_l) \tag{2}$$

where σ represents a non-linear activation function and ϕ denotes the kernel parameterized using neural network and W represents the bias term.

c) Projection Operation ($\mathcal{Q}: \mathbb{R}^{v_t} \to \mathbb{R}^{d_u}$): projects high-dimensional latent space to output using fully dense pointwise neural network along the channel dimension,

Overall, $G_\theta$ can be represented as follows:
$$G_\theta = \mathcal{Q} \circ K_l \circ \ldots \circ K_1 \circ \mathcal{P} \tag{3}$$

where ∘ denotes the composition operation.

Kernel Integral Operator: For the iterative updates in neural operators, the linear transformation is represented by a nonlocal kernel integral operator. It can be characterized using a kernel function κ:
$$K(a;\phi)v_l(x) = \int_D \kappa(x, y, a(x), a(y); \phi) v_l(y)\, dy \tag{4}$$

This kernel integral operation is formulated as a convolution operation which becomes multiplication operation in Fourier Domain. Let $\mathcal{F}$ represent the Fourier transform:
$$K(a;\phi)v_l(x) = \mathcal{F}^{-1}(\mathcal{F}(\kappa_\phi) \cdot \mathcal{F}(v_l))(x) \tag{5}$$

Hence, the FNO model parameterize the Fourier transform of this kernel function directly in Fourier space and the outputs are again converted to spatial domain as follows:
$$K(a;\phi)v_l(x) = \mathcal{F}^{-1}(R_\phi \cdot \mathcal{F}(v_l))(x) \tag{6}$$

In CoNOAir, the kernel integral is formulated using the Fractional Fourier Transform (FrFT) with a learnable order. The kernel integral is carried out within the complex-valued domain using Complex-Valued Neural Networks (CVNNs). Unlike, the FNO employs Fourier Transform to handle the integral kernel with real-valued inputs and outputs. In CoNOAir, the integral kernel can be mathematically formulated as:
$$K(a;\phi)v_l(x) = \mathcal{F}^{-\alpha}(R_\phi \cdot \mathcal{F}^{\alpha}(v_l))(x) \tag{7}$$

where $\mathcal{F}^{\alpha}$ denotes the order α fractional Fourier transform and $\mathcal{F}^{-\alpha}$ denotes inverse fractional Fourier transform. The kernel is parameterized as a learnable weight $R_\phi \in \theta$. In CoNOAir, α also becomes a learnable parameter.

*Model Training and Hyperparameters*

In this study, we used the variation of FNO, which performs convolution in the 2D spatial domain only and the input is fed into the model as a time series, just like a Recurrent Neural Network. A similar architecture for CoNOAir was used to compare their ability to make forecasts. The number of Fourier modes used was 12, with 20 hidden channels for each transformation unit. All the variations of these models were run for 500 epochs with an initial learning rate of 0.001 which was halved after every 100 epochs. $RL^2$ Norm, as given in Eq. 8 was used as the loss function with Adam as the optimizer.

$$RL^2 \ Norm = \frac{\sqrt{\sum_{u=1}^{n \times m}(Y_u - \widehat{Y_u})^2}}{\sqrt{\sum_{u=1}^{n \times m}(Y_u)^2}} \quad (8)$$

$Y_u$ is Ground Truth (WRF-Chem) and $\widehat{Y_u}$ is Predicted Value (FNO or CoNOAir). $n \times m$ represents the total number of grids after converting it into a 1D vector.

For our ablation studies we used 8, 10, 12 and 16 previous timesteps as inputs to our models, from which we concluded that 10 and 12 configurations worked the best. Finally, we selected the configuration of utilizing 10 previous timesteps as inputs to predict one timestep at a time. Subsequently, we can use the forward pass in an autoregressive fashion to output the required number of future steps. To update the models' parameters during the backpropagation of loss, we calculated the cumulative loss over the autoregressive forecasting for 4, 6 and 16 timesteps. The rationale behind this step was to develop models for both short-term and long-term forecasting by accumulating losses over smaller and longer time frames and to compare their abilities to make forecasts on both timescales. We established a uniform nomenclature for the rest of the paper to identify the different configurations of the models trained in this study. The name will be a combination of the type of architecture employed, whether it is CoNOAir or FNO and the number of timesteps (4, 6, or 16) used to propagate the loss backwards. For example, FNO(6) will represent the FNO architecture where 6 step forecasts were used to propagate the loss backwards and so on.

*Model Evaluation Metrics*

To evaluate our models' performance over short-term forecasting as well as long-term forecasting we have chosen three commonly used metrics namely Root Mean Squared Error (RMSE), Mean Absolute Error (MAE) and $RL^2$ Norm. We calculated the RMSE, MAE and $RL^2$ Norm between every predicted and corresponding observed image pair over all the samples and compared the mean and standard deviations. To calculate every metric, we converted the 2D arrays into 1D vectors and performed the respective operations. RMSE between one observed and predicted image was calculated as follows

$$\text{RMSE} = \sqrt{\frac{1}{n \times m} \times \sum_{u=1}^{n \times m}(Y_u - \widehat{Y_u})^2} \quad (9)$$

Similarly, MAE and $RL^2$ Norm were calculated as shown in Eqs. 10 and 8 respectively.

$$MAE = \frac{1}{n \times m} \times \sum_{u=1}^{n \times m} |Y_u - \widehat{Y_u}| \quad (10)$$

where RMSE is Root Mean Squared Error, MAE is Mean Absolute Error, $Y_u$ is Ground Truth (WRF-Chem) and $\widehat{Y_u}$ is Predicted Value (FNO or CoNOAir). $n \times m$ represents the total number of grids after converting it into a 1D vector.

In this study, we have evaluated our models' performance on the test data that was used for the validation during the training process, which comes from a similar distribution as the training data. Subsequently, we tested our models on an out of sample data which comes from the observed data of the 2019 year. For the test data, we forecasted in an autoregressive fashion for the next 16 hours and obtained the mean and standard deviation of the metrics defined above corresponding to every timestep's forecast. Similarly, for the whole out of sample data we have made forecasts for the next 20 hours. We also took 168 samples spanning over an entire week from the month of February 2019 to evaluate our models' capacity to forecast longer timeframes, i.e., the next 72 hours over the entire country whose results are shown in Table 1, Figure S2. We further analysed the short-term forecasting ability, i.e. the next one hour, of our models on the grids representing the six cities listed previously, along with the maximum concentration grid point. For this evaluation we have used the $R^2$ metric, RMSE and MAE to compare our model performance. The complete results are shown in Supplementary Material in Figure 3, Figure S4, Tables S2 and S3.

We also analysed the comparison between the time-series evolution for 72 hours using autoregressive forecasting with observed time series for these 7 grid locations, whose results are shown in Figure 5 and Table S4. Finally, we assessed our models' performance for the worst CO concentration events obtained using cumulative maximum concentration for the three crucial hours (i) morning rush, (ii) evening rush and (iii) midnight inversion illustrated in Figure 2 and Figure S3.

**References**

1    UN-Habitat. Envisaging the Future of Cities, World Cities Report 2022. (2022).
2    Jaffe, L. S. Ambient Carbon Monoxide And Its Fate in the Atmosphere. *Journal of the Air Pollution Control Association* **18**, 534-540, doi:10.1080/00022470.1968.10469168 (1968).
3    Gokhale, S. & Khare, M. A theoretical framework for the episodic-urban air quality management plan (e-UAQMP). *Atmospheric Environment* **41**, 7887-7894, doi:https://doi.org/10.1016/j.atmosenv.2007.06.061 (2007).
4    Aneja, V. P. *et al.* Measurements and analysis of criteria pollutants in New Delhi, India. *Environment International* **27**, 35-42, doi:https://doi.org/10.1016/S0160-4120(01)00051-4 (2001).
5    Kwiecień, J. & Szopińska, K. Mapping Carbon Monoxide Pollution of Residential Areas in a Polish City.  **12**, 2885 (2020).
6    Liñán-Abanto, R. N. *et al.* Temporal variations of black carbon, carbon monoxide, and carbon dioxide in Mexico City: Mutual correlations and evaluation of emissions inventories. *Urban Climate* **37**, 100855, doi:https://doi.org/10.1016/j.uclim.2021.100855 (2021).
7    Pani, S. K. *et al.* Relationship between long-range transported atmospheric black carbon and carbon monoxide at a high-altitude background station in East Asia. *Atmospheric Environment* **210**, 86-99, doi:https://doi.org/10.1016/j.atmosenv.2019.04.053 (2019).
8    Novelli, P. C., Masarie, K. A., Tans, P. P. & Lang, P. M. Recent changes in atmospheric carbon monoxide. *Science* **263**, 1587-1590, doi:10.1126/science.263.5153.1587 (1994).
9    Holloway, T., Levy, H. & Kasibhatla, P. Global distribution of carbon monoxide. *Journal of Geophysical Research: Atmospheres* **105**, 12123-12147, doi:10.1029/1999jd901173 (2000).
10   Calderon-Garciduenas, L. *et al.* Air pollution, cognitive deficits and brain abnormalities: a pilot study with children and dogs. *Brain Cogn* **68**, 117-127, doi:10.1016/j.bandc.2008.04.008 (2008).
11   Levy, R. J. Carbon monoxide pollution and neurodevelopment: A public health concern. *Neurotoxicol Teratol* **49**, 31-40, doi:10.1016/j.ntt.2015.03.001 (2015).
12   Liu, C. *et al.* Ambient carbon monoxide and cardiovascular mortality: a nationwide time-series analysis in 272 cities in China. *Lancet Planet Health* **2**, e12-e18, doi:10.1016/S2542-5196(17)30181-X (2018).

13   Satran, D. *et al.* Cardiovascular manifestations of moderate to severe carbon monoxide poisoning. *J Am Coll Cardiol* **45**, 1513-1516, doi:10.1016/j.jacc.2005.01.044 (2005).
14   Kura, B., Verma, S., Ajdari, E. & Iyer, A. Growing Public Health Concerns from Poor Urban Air Quality: Strategies for Sustainable Urban Living %J Computational Water, Energy, and Environmental Engineering.  **Vol.02No.02**, 9, doi:10.4236/cweee.2013.22B001 (2013).
15   Gulia, S., Shiva Nagendra, S. M., Khare, M. & Khanna, I. Urban air quality management-A review. *Atmospheric Pollution Research* **6**, 286-304, doi:https://doi.org/10.5094/APR.2015.033 (2015).
16   Zhang, X. *et al.* Linking urbanization and air quality together: A review and a perspective on the future sustainable urban development. *Journal of Cleaner Production* **346**, 130988, doi:https://doi.org/10.1016/j.jclepro.2022.130988 (2022).
17   CPCB. *Central Control Room for Air Quality Management - All India*, <https://airquality.cpcb.gov.in/ccr/#/caaqm-dashboard-all/caaqm-landing> (
18   Brauer, M. *et al.* Examination of monitoring approaches for ambient air pollution: A case study for India. *Atmospheric Environment* **216**, doi:10.1016/j.atmosenv.2019.116940 (2019).
19   Guttikunda, S., Ka, N., Ganguly, T. & Jawahar, P. Plugging the ambient air monitoring gaps in India's national clean air programme (NCAP) airsheds. *Atmospheric Environment* **301**, 119712, doi:https://doi.org/10.1016/j.atmosenv.2023.119712 (2023).
20   Chandra, N., Venkataramani, S., Lal, S., Sheel, V. & Pozzer, A. Effects of convection and long-range transport on the distribution of carbon monoxide in the troposphere over India. *Atmospheric Pollution Research* **7**, 775-785, doi:10.1016/j.apr.2016.03.005 (2016).
21   Yarragunta, Y., Srivastava, S. & Mitra, D. Validation of lower tropospheric carbon monoxide inferred from MOZART model simulation over India. *Atmospheric Research* **184**, 35-47, doi:10.1016/j.atmosres.2016.09.010 (2017).
22   Elguindi, N. *et al.* Current status of the ability of the GEMS/MACC models to reproduce the tropospheric CO vertical distribution as measured by MOZAIC. *Geoscientific Model Development* **3**, 501-518, doi:10.5194/gmd-3-501-2010 (2010).
23   Surendran, D. E. *et al.* Air quality simulation over South Asia using Hemispheric Transport of Air Pollution version-2 (HTAP-v2) emission inventory and Model for Ozone and Related chemical Tracers (MOZART-4). *Atmospheric Environment* **122**, 357-372, doi:10.1016/j.atmosenv.2015.08.023 (2015).
24   I, N., Srivastava, S., Yarragunta, Y., Kumar, R. & Mitra, D. Distribution of surface carbon monoxide over the Indian subcontinent: Investigation of source contributions using WRF-Chem. *Atmospheric Environment* **243**, doi:10.1016/j.atmosenv.2020.117838 (2020).
25   Kumar, R. *et al.* Simulations over South Asia using the Weather Research and Forecasting model with Chemistry (WRF-Chem): chemistry evaluation and initial results. *Geoscientific Model Development* **5**, 619-648, doi:10.5194/gmd-5-619-2012 (2012).
26   Tie, X. *et al.* Characterizations of chemical oxidants in Mexico City: A regional chemical dynamical model (WRF-Chem) study. *Atmospheric Environment* **41**, 1989-2008, doi:https://doi.org/10.1016/j.atmosenv.2006.10.053 (2007).
27   Benavente, N. R. *et al.* Air quality simulation with WRF-Chem over southeastern Brazil, part I: Model description and evaluation using ground-based and satellite data. *Urban Climate* **52**, 101703, doi:https://doi.org/10.1016/j.uclim.2023.101703 (2023).
28   Carmichael, G. R. *et al.* Predicting air quality: Improvements through advanced methods to integrate models and measurements. *Journal of Computational Physics* **227**, 3540-3571, doi:10.1016/j.jcp.2007.02.024 (2008).
29   Werhahn, M., Xie, Y., Chu, M. & Thuerey, N. A Multi-Pass GAN for Fluid Flow Super-Resolution. *Proceedings of the ACM on Computer Graphics and Interactive Techniques* **2**, 1-21, doi:10.1145/3340251 (2019).

30  Almubaidin, M. A. *et al.* Machine learning predictions for carbon monoxide levels in urban environments. *Results in Engineering* **22**, doi:10.1016/j.rineng.2024.102114 (2024).
31  Wong, P.-Y. *et al.* Incorporating land-use regression into machine learning algorithms in estimating the spatial-temporal variation of carbon monoxide in Taiwan. *Environmental Modelling & Software* **139**, doi:10.1016/j.envsoft.2021.104996 (2021).
32  Feizi, H., Sattari, M. T., Prasad, R. & Apaydin, H. Comparative analysis of deep and machine learning approaches for daily carbon monoxide pollutant concentration estimation. *International Journal of Environmental Science and Technology* **20**, 1753-1768, doi:10.1007/s13762-022-04702-x (2023).
33  Bougoudis, I., Demertzis, K. & Iliadis, L. HISYCOL a hybrid computational intelligence system for combined machine learning: the case of air pollution modeling in Athens. *Neural Computing and Applications* **27**, 1191-1206, doi:10.1007/s00521-015-1927-7 (2015).
34  Suárez Sánchez, A., García Nieto, P. J., Riesgo Fernández, P., del Coz Díaz, J. J. & Iglesias-Rodríguez, F. J. Application of an SVM-based regression model to the air quality study at local scale in the Avilés urban area (Spain). *Mathematical and Computer Modelling* **54**, 1453-1466, doi:10.1016/j.mcm.2011.04.017 (2011).
35  Barthwal, A. & Goel, A. K. Advancing air quality prediction models in urban India: a deep learning approach integrating DCNN and LSTM architectures for AQI time-series classification. *Modeling Earth Systems and Environment* **10**, 2935-2955, doi:10.1007/s40808-023-01934-9 (2024).
36  Moazami, S. *et al.* Reliable prediction of carbon monoxide using developed support vector machine. *Atmospheric Pollution Research* **7**, 412-418, doi:10.1016/j.apr.2015.10.022 (2016).
37  Navares, R. & Aznarte, J. L. Predicting air quality with deep learning LSTM: Towards comprehensive models. *Ecological Informatics* **55**, doi:10.1016/j.ecoinf.2019.101019 (2020).
38  Xu, S., Li, W., Zhu, Y. & Xu, A. A novel hybrid model for six main pollutant concentrations forecasting based on improved LSTM neural networks. *Scientific Reports* **12**, 14434, doi:10.1038/s41598-022-17754-3 (2022).
39  Alimissis, A., Philippopoulos, K., Tzanis, C. G. & Deligiorgi, D. Spatial estimation of urban air pollution with the use of artificial neural network models. *Atmospheric Environment* **191**, 205-213, doi:https://doi.org/10.1016/j.atmosenv.2018.07.058 (2018).
40  Li, C. & Zhu, Z. Research and application of a novel hybrid air quality early-warning system: A case study in China. *Science of The Total Environment* **626**, 1421-1438, doi:https://doi.org/10.1016/j.scitotenv.2018.01.195 (2018).
41  Kovachki, N. *et al.* Neural Operator: Learning Maps Between Function Spaces. arXiv:2108.08481 (2021). <https://ui.adsabs.harvard.edu/abs/2021arXiv210808481K>.
42  Li, Z. *et al.* Fourier neural operator for parametric partial differential equations.  (2020).
43  Tiwari, K., Krishnan, N. & Prathosh, A. J. a. p. a. CoNO: Complex Neural Operator for Continous Dynamical Physical Systems.  (2024).
44  Guttikunda, S., Ka, N., Ganguly, T. & Jawahar, P. Plugging the ambient air monitoring gaps in India's national clean air programme (NCAP) airsheds. *Atmospheric Environment* **301**, doi:10.1016/j.atmosenv.2023.119712 (2023).
45  CPCB.   (ed Central Pollution Control Board) (New Delhi, India, 2014).
46  Sharma, S., Chandra, M. & Harsha Kota, S. Four year long simulation of carbonaceous aerosols in India: Seasonality, sources and associated health effects. *Environ Res* **213**, 113676, doi:10.1016/j.envres.2022.113676 (2022).
47  Nayak, D. K., Habib, G. & Kota, S. H. Can Landuse Landcover changes influence the success of India's national clean air plans? *Atmospheric Environment: X* **22**, 100251 (2024).

48  Gupta, M., Nayak, D. K. & Harsha Kota, S. Impact of particulate matter-centric clean air action plans on ozone concentrations in India. *ACS Earth and Space Chemistry* **7**, 1038-1048 (2023).

49  Jat, R., Gurjar, B. R. & Lowe, D. Regional pollution loading in winter months over India using high resolution WRF-Chem simulation. *Atmospheric Research* **249**, doi:10.1016/j.atmosres.2020.105326 (2021).

50  Ritter, M., Müller, M. D., Jorba, O., Parlow, E. & Liu, L. J. S. Impact of chemical and meteorological boundary and initial conditions on air quality modeling: WRF-Chem sensitivity evaluation for a European domain. *Meteorology and Atmospheric Physics* **119**, 59-70, doi:10.1007/s00703-012-0222-8 (2012).

51  Burark, P., Tiwari, K., Rashid, M. M., A. P, P. & Krishnan, N. M. A. CoDBench: a critical evaluation of data-driven models for continuous dynamical systems. *Digital Discovery* **3**, 1172-1181, doi:10.1039/d4dd00028e (2024).

52  Takamoto, M. *et al.* Pdebench: An extensive benchmark for scientific machine learning. **35**, 1596-1611 (2022).

**Data and Codes**

Codes and Data for training and evaluation of the models will be shared after the acceptance of the manuscript.

**Supplementary Material**

*S.1. Evaluation of Loss Curves and Training Process*

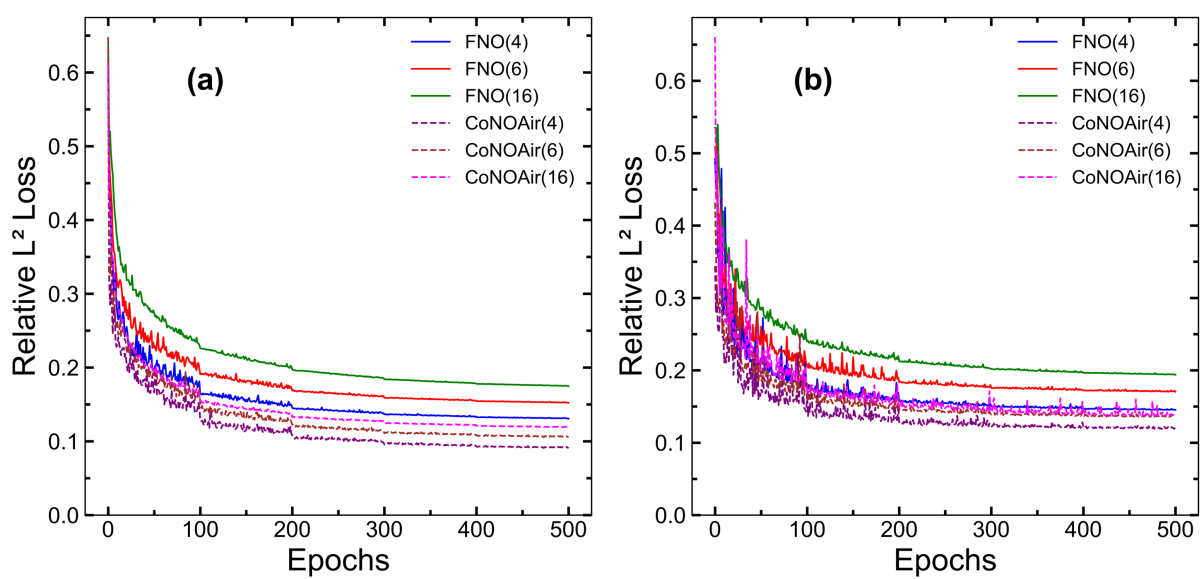

**Figure S1. Train (a) and test (b) loss curves for FNO and CoNOAir models trained in this study.**

Evaluation of every ML model begins with the investigation of the loss curve obtained during training to ensure that the model learns the necessary information from the training data. The loss curve must follow a downward trend over the number of epochs for the training dataset. To ensure that the model not only performs well for the training data, but also on the unseen data, the loss is calculated on the unseen test data for every epoch to visualise its trend. Figure S1 (a) indicates that as training progressed, all six models showed a decrease in training loss, implying that they learnt the underlying relationship between input and output functions efficiently. Subsequently, it is noticed that the final training loss for the CoNOAir(4) model is the lowest, while FNO(16) has the highest at the end of 500 epochs. All the variations of CoNOAir have a final training loss smaller than that of the best possible FNO variation. The final training loss increases with an increase in the number of timesteps over which the loss is accumulated, i.e., the final loss for the variation with 4 timesteps is less than that of 6 timesteps, which itself is less than the 16 timesteps variation for both the architectures. This was anticipated because with more timesteps, it becomes increasingly difficult to converge the accumulated loss beyond a certain point. A similar trend can be seen in Figure S1 (b) for the test loss curves, where the CoNOAir variations performed better than the corresponding FNO variations. The worst CoNOAir variation performed better than all the FNO variations, and the final test loss for CoNOAir(6) and CoNOAir(16) are very close to each other, with CoNOAir(16) having a slightly higher test loss.

*S.2. Autoregressive loss over rollout*

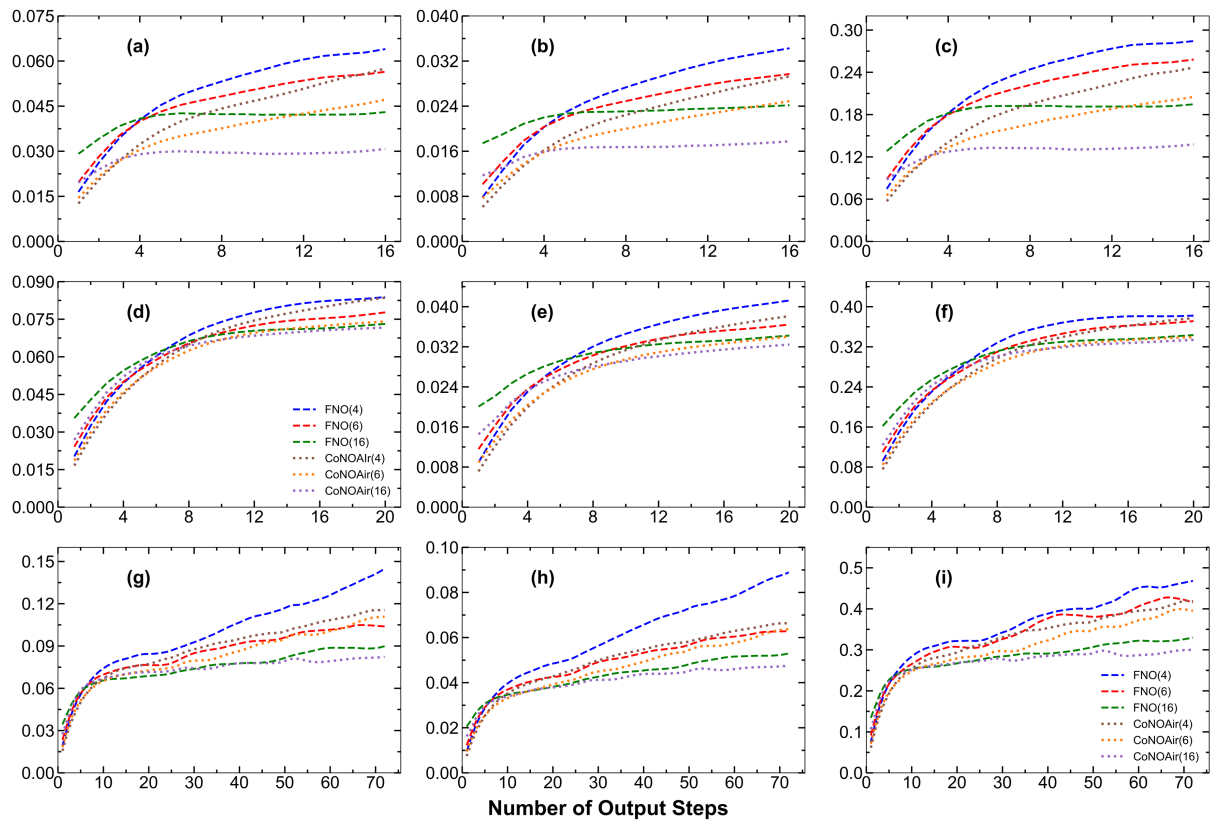

**Figure S2. Evolution of error metrics over validation and test data:** *Three error metrics have been used, namely RMSE, MAE and RL$^2$ Norm on three different datasets as validation set used during the training of models, whole out of sample test dataset, subset of test dataset representing higher concentration days. (a), (d) and (g) represent the RMSE errors on these three datasets respectively. (b), (e) and (h) represent the MAE errors, while (c), (f) and (i) represent the RL$^2$ norm error for these datasets, respectively.*

Table S1 and Figure S2 (a) illustrate that mean RMSE ± standard deviation is the lowest for CoNOAir(4) at 0.013 ± 0.008 followed by CoNOAir(6) at 0.014 ± 0.008 and FNO(4) at 0.016 ± 0.01 for the first timestep prediction. FNO(16) has the highest mean RMSE until 4 hours of forecasting, beyond which FNO(4) followed by FNO(6) have the highest RMSE. CoNOAir(4) and CoNOAir(6) also perform poorly as compared to the CoNOAir(16) beyond 3-4 hours of forecasting. The steepest increase in the errors is seen in the (4) variations for both the architectures, with CoNOAir(4)(0.058±0.036) having a lower RMSE than FNO(4)(0.064±0.041) for 16th-hour forecasts. The CoNOAir(16) with 0.031±0.017 has the lowest RMSE followed by FNO(16) at 0.043±0.022 and CoNOAir(6) at 0.047±0.028 in the respective order for the 16$^{th}$ hour forecasts. From Figure S2 (b), we can observe that for MAE error a similar trend is observed except for the FNO(6) having MAE slightly less than CoNOAir(16) for 1 hour's forecasting. For the 16$^{th}$ hour forecasting, CoNOAir(16) has the lowest MAE at 0.018±0.009, followed by FNO(16) at 0.024±0.012. The CoNOAir(6)'s MAE (0.025±0.014) is comparable to FNO(16)'s MAE (0.024±0.012); meanwhile, CoNOAir(4) at 0.029±0.017 and FNO(6) at 0.03±0.016 have comparable MAE at 16$^{th}$ hour. FNO(4) has the worst RMSE and MAE errors for 16$^{th}$ hour forecasting. The same configurations for CoNOAir and FNO have a kind of parallel curve for the evolution of error over the timesteps for both MAE and RMSE. A very similar trend is observed in Figure S2 (c) and Table 1, for RL$^2$ Norm error's evolution for the test dataset, with CoNOAir(16) performing the best at 0.138±0.042 and FNO(4) performing the worst (0.284±0.099) for the 16$^{th}$ hour forecasting. Meanwhile, the CoNOAir(4) (0.057±0.026) has the lowest and FNO(16) (0.128±0.034) the highest RL$^2$ Norm for 1$^{st}$ hour forecasting.

Figure S2 (d), (e), (f) show the RMSE, MAE, and RL$^2$ norm evolution over 20 timesteps, forecasting autoregressively for the out of sample data respectively. From Table S1 and Figure S2 (d) we conclude that CoNOAir(4) configuration has the lowest RMSE (0.017±0.009) for 1$^{st}$ hour forecasting followed by CoNOAir(6) at 0.019±0.009 and FNO(4) at 0.02±0.01 respectively. FNO(16) has the highest RMSE (0.036±0.015) among all the variations for 1$^{st}$ hour forecasting. For 20$^{th}$ hour forecasting, the CoNOAir(16) variation has the lowest RMSE (0.072±0.03), succeeded by FNO(16) at 0.073±0.031. For 20$^{th}$ hour forecasts, CoNOAir(6) has the third least RMSE (0.074±0.031); FNO(4) (0.084±0.046) and CoNOAir(4) (0.084±0.039) have close RMSE values, despite CoNOAir(4) having the smaller error at the initial timesteps. A similar trend is followed for MAE, where CoNOAir(4) performs the best for 1$^{st}$ hour (0.007±0.003) and CoNOAir(16) performs the best for 20$^{th}$ hour forecast (0.032±0.009). CoNOAir(6)'s MAE (0.034±0.009) is equal to that of FNO(16) for the 20$^{th}$ hour. For the RL$^2$ norm, a resembling pattern is observed, where CoNOAir(4) performs the best (0.075±0.036) for the next hour forecasting, and CoNOAir(16) performs the best (0.334±0.125) for the 20$^{th}$ hour. For all the three errors, FNO(4) and CoNOAir(4) models have a swift development of errors over time. FNO(16) and CoNOAir(6) having similar errors for the long-term forecasting indicates that the CoNOAir(6) model is able to learn enough information as FNO(16) from the provided data.

Figure S2 (g), (h), (i) depict the development of three errors respectively, for 72 hours forecasting over the 168 samples taken from the month of February 2019. RMSE at the 1$^{st}$ hour forecasting has a similar trend as observed in Figure S2 (a), (b), (c), (d), (e) and (f). For the 72th hour forecasting, RMSE is lowest for the CoNOAir(16) (0.082±0.024) followed by FNO(16) at 0.09±0.027. Interestingly, RMSE for CoNOAir(6) increases rapidly and FNO(6) variation has the third lowest RMSE (0.104±0.027) among all the models for forecasting 72$^{nd}$ hour. FNO(4) performs the worst (0.145±0.046) out of all the 6 models in terms of RMSE. An equivalent trend can be observed for MAE evolution where CoNOAir(16) at 0.047±0.01 and FNO(16) at 0.053±0.013 have the lowest errors respectively, followed by FNO(6) (0.063±0.012) and CoNOAir(6) (0.064±0.013) for forecasting 72$^{nd}$ hour. Both the latter models have a close MAE value at the 72$^{nd}$ hour forecasting, and FNO(4) has the worst performance at 72$^{nd}$ hour (0.089±0.02). CoNOAir(16) at 0.3±0.063 performs the best in terms of RL$^2$ Norm succeeded by FNO(16) at 0.33±0.055 for 72$^{nd}$ hour forecasting. CoNOAir(6) (0.395±0.15) has better performance than CoNOAir(4) (0.419±0.1) and FNO(6) (0.415±0.086) from the perspective of RL$^2$ norm, while the FNO(4) (0.468±0.073) performs the worst at 72$^{nd}$ hour. For forecasting 30-35 hours, CoNOAir(6) is giving sufficient performance as for 1 hour its performance is comparable to CoNOAir(4) and for 35$^{th}$ hour its comparable to CoNOAir(16) and FNO(16).

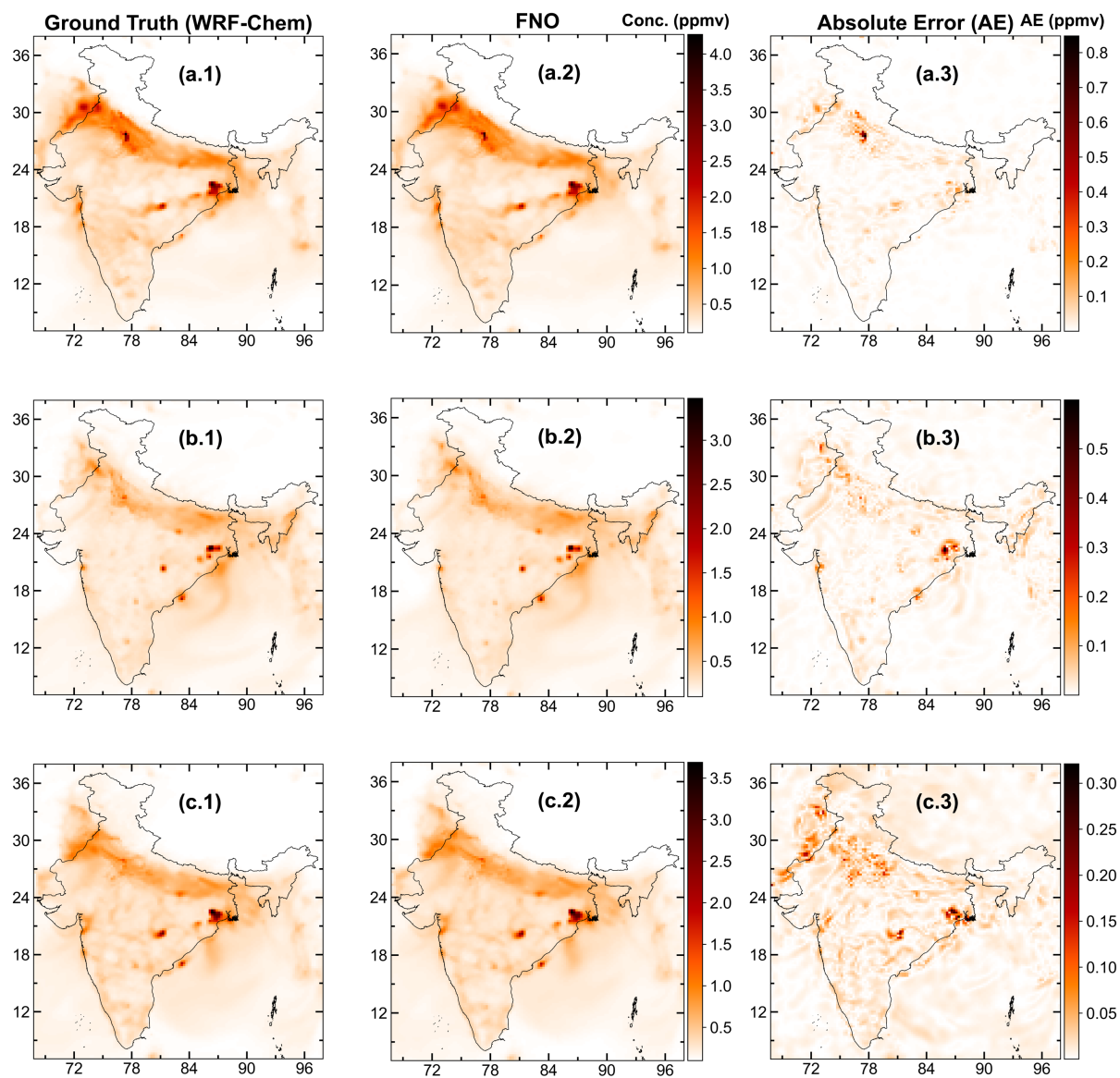

**Figure S3. Prediction of FNO(4) on the maximum concentration hours:** *Both the ground truth and FNO maps are normalized to the same range for the purpose of visualisation. Ground Truth for 12th February 2019 07:00 am IST (a.1), 6th February 19:00 IST (b.1) and 6th February 00:00 IST (c.1). CO concentrations predicted using FNO(4) for the corresponding hours (a.2), (b.2) and (c.2). Absolute error between the ground truth and FNO(4) predictions (a.3), (b.3) and (c.3).*

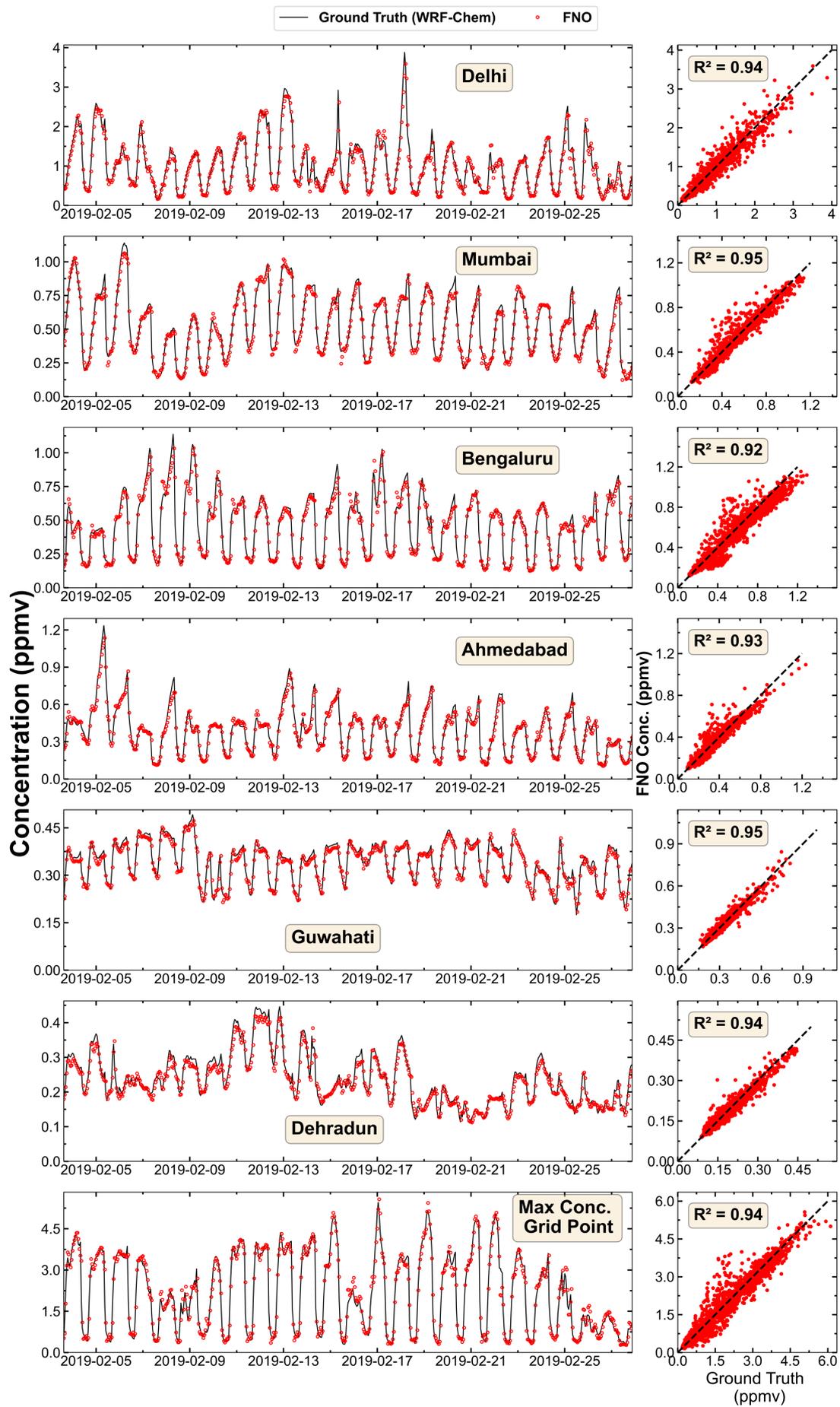

**Figure S4. Short-term forecast for cities.** *Time series 1-hour forecast for FNO(4) for 6 cities and maximum concentration grid point to evaluate the ability of FNO for grid-level short-term forecasts. A Time-series comparison is shown for February month in the left column. The column on the right shows the predicted values with respect to the ground truth concentrations for the whole test dataset. The $R^2$ values corresponding to each grid point are shown in the inset.*

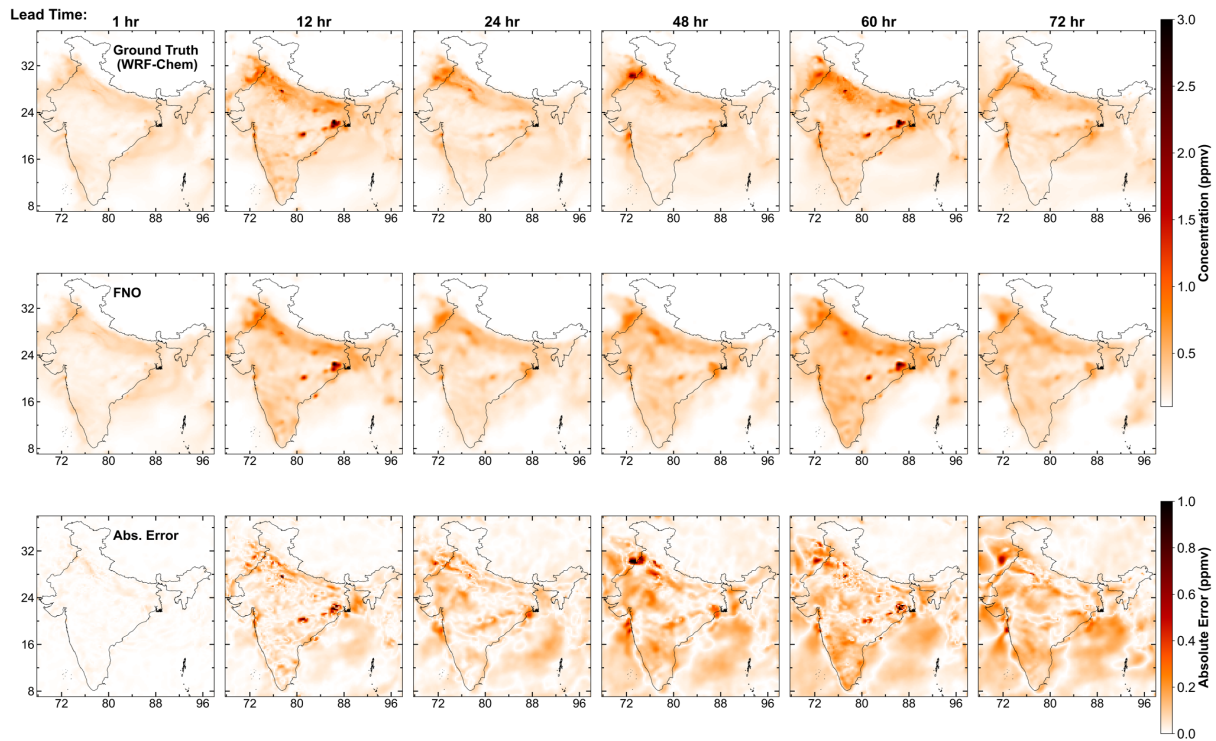

**Figure S5. Country-wide evolution of long-term forecast:** *72 hours autoregressive forecasts for the same instance as Figure 4 for the whole country for evaluating the long-term forecasting using FNO(4) model.*

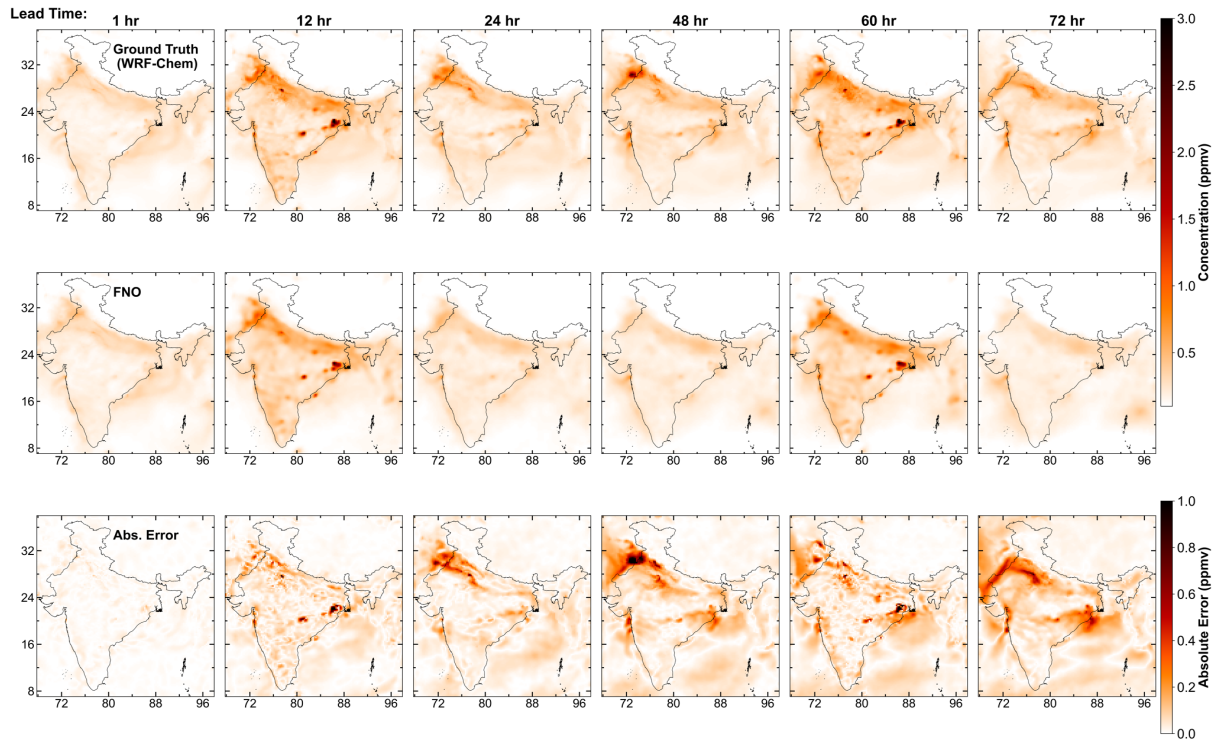

**Figure S6. Country-wide evolution of long-term forecast:** *72 hours autoregressive forecasts for the same instance as Figure 4 for the whole country for evaluating the long-term forecasting using FNO(6) model.*

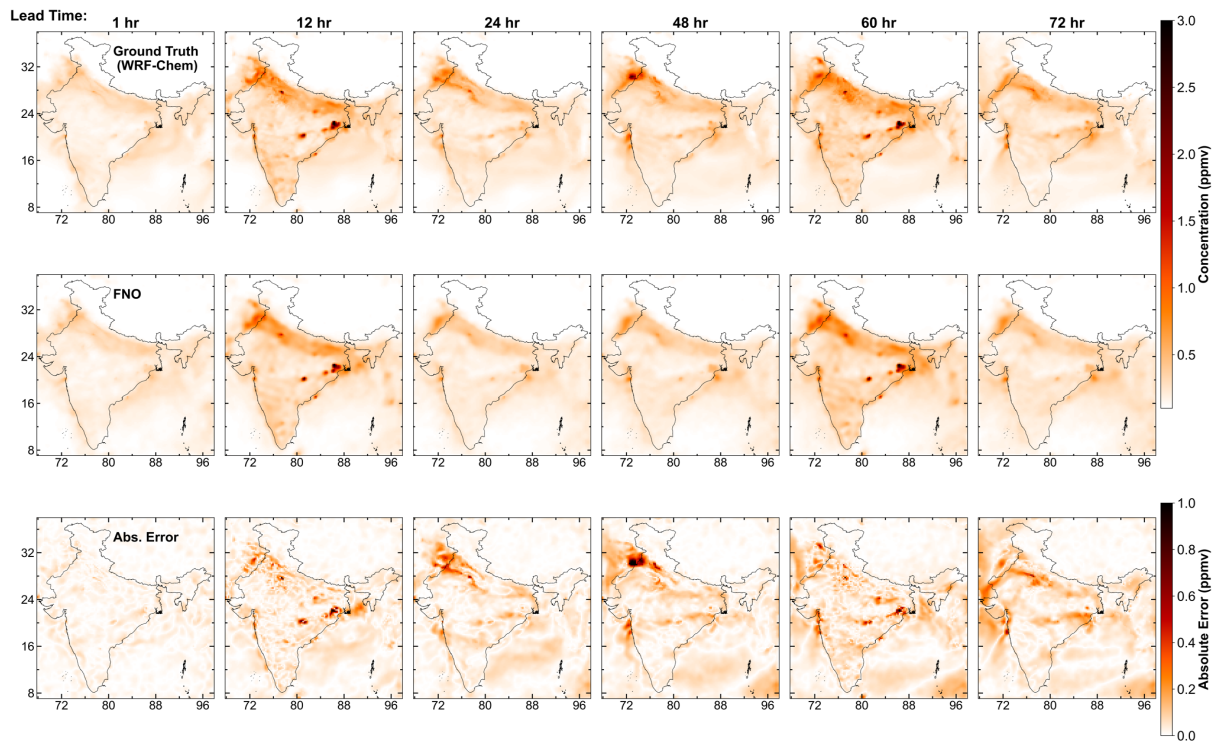

**Figure S7. Country-wide evolution of long-term forecast.** *72 hours autoregressive forecasts for the same instance as Figure 4 for the whole country for evaluating the long-term forecasting using FNO(16) model.*

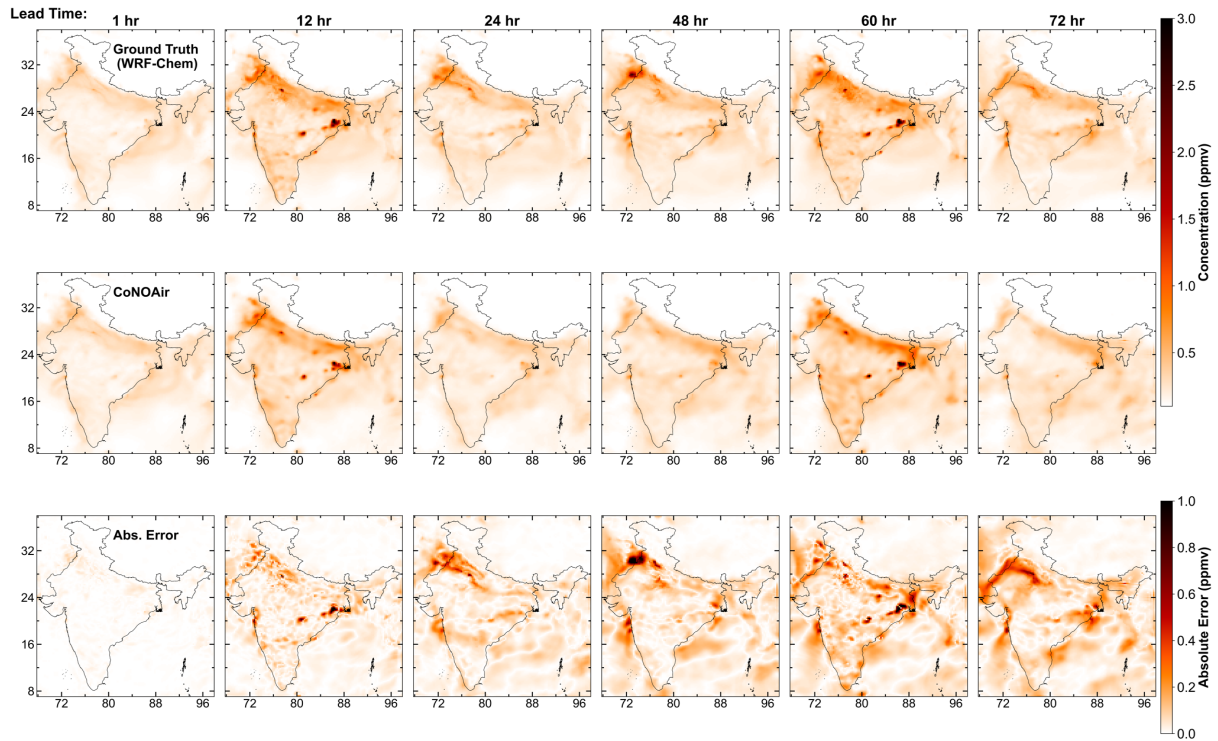

**Figure S8. Country-wide evolution of long-term forecast:** *72 hours autoregressive forecasts for the same instance as Figure 4 for the whole country for evaluating the long-term forecasting capability using CoNOAir(4) model.*

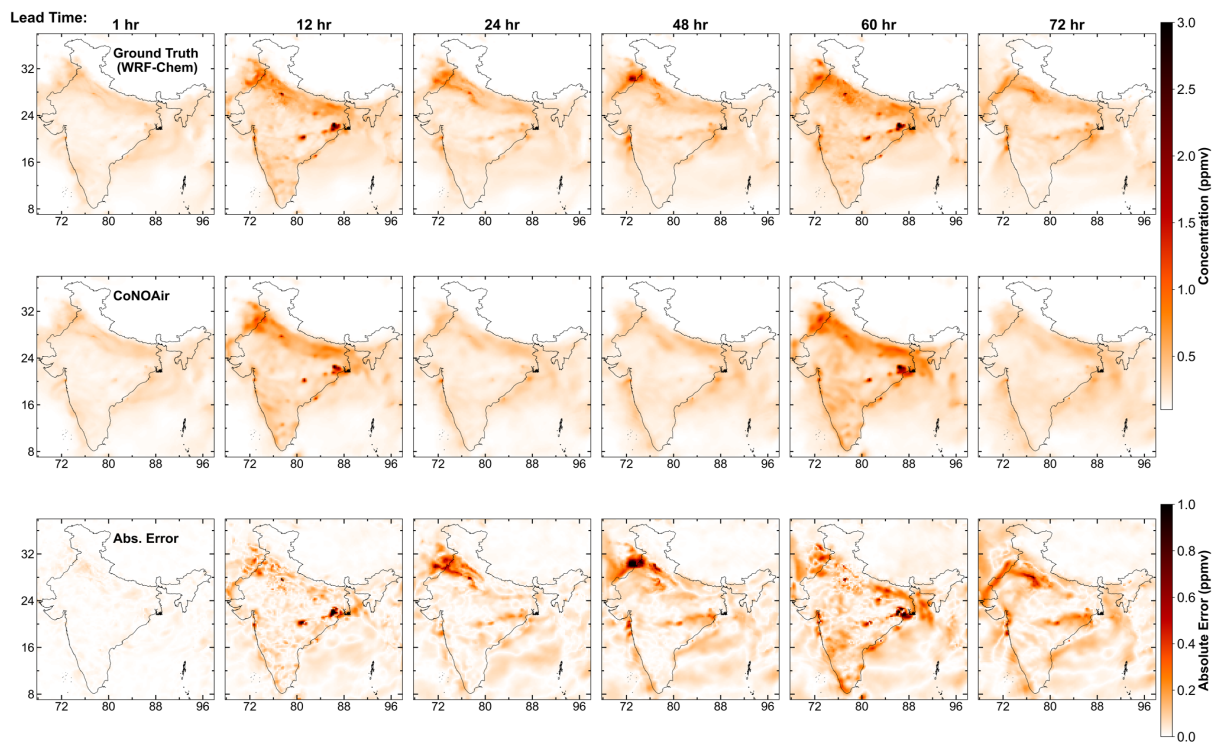

**Figure S9. Country-wide evolution of long-term forecast:** *72 hours autoregressive forecasts for the same instance as Figure 4 for the whole country for evaluating the long-term forecasting capability using CoNOAir(6) model.*

**Table S1: Evaluation of 1 hour forecasting capability of various models for validation set used during the training of models, whole out of sample test dataset, and subset of test dataset representing higher concentration days.**

| Error → <br> Model ↓ | FNO(4) | FNO(6) | FNO(16) | CoNOAir(4) | CoNOAir(6) | CoNOAir(16) |
|---|---|---|---|---|---|---|
| Validation Data (Errors for 1st hour forecasts) | | | | | | |
| RMSE | 0.016 (0.01) | 0.02 (0.01) | 0.029 (0.014) | 0.013 (0.008) | 0.014 (0.008) | 0.02 (0.009) |
| MAE | 0.008 (0.004) | 0.01 (0.005) | 0.017 (0.008) | 0.006 (0.003) | 0.008 (0.003) | 0.012 (0.005) |
| RL$^2$ Norm | 0.075 (0.032) | 0.088 (0.032) | 0.128 (0.034) | 0.057 (0.026) | 0.065 (0.025) | 0.088 (0.025) |
| Test Data (Errors for 1st hour forecasts) | | | | | | |
| RMSE | 0.02 (0.01) | 0.024 (0.011) | 0.036 (0.015) | 0.017 (0.009) | 0.019 (0.009) | 0.027 (0.01) |
| MAE | 0.009 (0.003) | 0.012 (0.004) | 0.02 (0.008) | 0.007 (0.003) | 0.009 (0.003) | 0.015 (0.003) |
| RL$^2$ Norm | 0.092 (0.04) | 0.11 (0.041) | 0.162 (0.044) | 0.075 (0.036) | 0.085 (0.036) | 0.124 (0.037) |
| Test Data (Errors for 1st hour forecasts) | | | | | | |
| RMSE | 0.019 (0.008) | 0.023 (0.008) | 0.034 (0.009) | 0.016 (0.007) | 0.018 (0.007) | 0.027 (0.007) |
| MAE | 0.009 (0.003) | 0.012 (0.003) | 0.021 (0.004) | 0.007 (0.002) | 0.009 (0.002) | 0.016 (0.003) |
| RL$^2$ Norm | 0.075 (0.027) | 0.091 (0.026) | 0.134 (0.023) | 0.061 (0.025) | 0.071 (0.024) | 0.107 (0.023) |

**Table S2: RMSE for short-term forecasting of the ML models for grid level forecasting.**

| Model | FNO(4) | FNO(6) | FNO(16) | CoNOAir(4) | CoNOAir(6) | CoNOAir(16) |
|---|---|---|---|---|---|---|
| Delhi | 0.14 | 0.15 | 0.19 | 0.13 | 0.14 | 0.15 |
| Mumbai | 0.05 | 0.06 | 0.07 | 0.04 | 0.04 | 0.05 |
| Bengaluru | 0.08 | 0.08 | 0.1 | 0.06 | 0.06 | 0.07 |
| Ahmedabad | 0.04 | 0.05 | 0.06 | 0.03 | 0.03 | 0.04 |
| Guwahati | 0.02 | 0.03 | 0.04 | 0.02 | 0.02 | 0.03 |
| Dehradun | 0.02 | 0.02 | 0.03 | 0.01 | 0.02 | 0.02 |
| Max Conc. Grid Point | 0.32 | 0.35 | 0.42 | 0.25 | 0.28 | 0.31 |

**Table S3: MAE for short-term forecasting of the ML models for grid level forecasting.**

| Model | FNO(4) | FNO(6) | FNO(16) | CoNOAir(4) | CoNOAir(6) | CoNOAir(16) |
|---|---|---|---|---|---|---|
| Delhi | 0.09 | 0.1 | 0.12 | 0.08 | 0.08 | 0.1 |
| Mumbai | 0.03 | 0.04 | 0.05 | 0.03 | 0.03 | 0.04 |
| Bengaluru | 0.05 | 0.06 | 0.07 | 0.04 | 0.04 | 0.05 |
| Ahmedabad | 0.03 | 0.03 | 0.04 | 0.02 | 0.02 | 0.03 |
| Guwahati | 0.01 | 0.02 | 0.03 | 0.01 | 0.02 | 0.02 |
| Dehradun | 0.01 | 0.01 | 0.03 | 0.01 | 0.01 | 0.02 |
| Max Conc. Grid Point | 0.21 | 0.24 | 0.3 | 0.16 | 0.19 | 0.21 |

**Table S4: RMSE for 72 hours autoregressive forecasting for different cities and max conc. grid points using different models. The values represent the mean (standard deviation) for RMSE across samples.**

| City | FNO(4) | FNO(6) | FNO(16) | CoNOAir(4) | CoNOAir(6) | CoNOAir(16) |
|---|---|---|---|---|---|---|
| Delhi | 0.266 (0.142) | 0.313 (0.198) | 0.206 (0.122) | 0.353 (0.154) | 0.274 (0.164) | 0.218 (0.134) |
| Mumbai | 0.035 (0.023) | 0.029 (0.012) | 0.026 (0.026) | 0.037 (0.024) | 0.024 (0.018) | 0.028 (0.017) |
| Bengaluru | 0.026 (0.015) | 0.024 (0.013) | 0.027 (0.022) | 0.022 (0.011) | 0.026 (0.016) | 0.02 (0.011) |
| Ahmedabad | 0.028 (0.016) | 0.022 (0.015) | 0.017 (0.007) | 0.018 (0.013) | 0.021 (0.006) | 0.013 (0.007) |
| Guwahati | 0.018 (0.015) | 0.005 (0.004) | 0.003 (0.002) | 0.008 (0.006) | 0.006 (0.003) | 0.003 (0.003) |
| Dehradun | 0.013 (0.008) | 0.011 (0.008) | 0.01 (0.005) | 0.013 (0.008) | 0.008 (0.004) | 0.007 (0.004) |
| Max Conc. Grid Point | 0.961 (0.411) | 0.96 (0.403) | 0.614 (0.35) | 0.843 (0.584) | 0.895 (0.418) | 0.69 (0.513) |